\documentclass{article}

     \usepackage[preprint]{neurips_2018}

\usepackage[utf8]{inputenc} 
\usepackage[T1]{fontenc}    
\usepackage{hyperref}       
\usepackage{url}            
\usepackage{booktabs}       
\usepackage{amsfonts}       
\usepackage{nicefrac}       
\usepackage{microtype}      
\usepackage{subfig}
\usepackage{graphicx}
\usepackage{amsmath}

\title{LapTool-Net: A Contextual Detector of Surgical Tools in Laparoscopic Videos Based on Recurrent Convolutional Neural Networks}

\author{
  Babak Namazi\\
  Department of Electrical Engineering\\
  University of Texas at Arlington\\
  Arlington, TX 76019 \\
  \texttt{babak.namazi@mavs.uta.edu} \\
   \And
 Ganesh Sankaranarayanan \\
  Department of Surgery\\
  Baylor University Medical Center\\
  Dallas, TX 75246 \\
  \texttt{ganesh.sankaranarayanan@bswhealth.org} \\
   \AND
   Venkat Devarajan \\
   Department of Electrical Engineering\\
  University of Texas at Arlington\\
  Arlington, TX 76019 \\
   \texttt{venkat@uta.edu} \\
}

\begin{document}
\maketitle

\begin{abstract}
We propose a new multilabel classifier, called LapTool-Net to detect the presence of surgical tools in each frame of a laparoscopic video. The novelty of LapTool-Net is the exploitation of the correlations among the usage of different tools and, the tools and tasks - i.e., the context of the tools' usage. Towards this goal, the pattern in the co-occurrence of the tools is utilized for designing a decision policy for a multilabel classifier based on a Recurrent Convolutional Neural Network (RCNN) architecture to simultaneously extract the spatio-temporal features. In contrast to the previous multilabel classification methods, the RCNN and the decision model are trained in an end-to-end manner using a multi-task learning scheme. To overcome the high imbalance and avoid overfitting caused by the lack of variety in the training data, a high down-sampling rate is chosen based on the more frequent combinations. Furthermore, at the post-processing step, the predictions for all the frames of a video are corrected by designing a bi-directional RNN to model the long-term tasks’ order. LapTool-Net was trained using a publicly available dataset of laparoscopic cholecystectomy. The results show LapTool-Net outperformed existing methods significantly, even while using fewer training samples and a shallower architecture.
\end{abstract}


\section{Introduction}
\label{sec1}
Numerous advantages of minimally invasive surgery such as shorter recovery time, less pain and blood loss, and better cosmetic results, make it the preferred choice over conventional open surgeries \citep{Velanovich2000}. In laparoscopy, the surgical instruments are inserted through small incisions in the abdominal wall and the procedure is monitored using a laparoscope. The special way of manipulating the surgical instruments and the indirect observation of the surgical scene introduce more challenges in performing laparoscopic procedures \citep{Ballantyne2002TheSurgery.}. The complexity of laparoscopy requires special training and assessment for the surgery residents to gain the required bi-manual dexterity. The videos from the previously accomplished procedures by expert surgeons can be used for such training and assessment. The tedium and cost of such an assessment can be dramatically reduced using an automated tool detection system, among other things and is, therefore, the focus of this paper.

In computer-aided intervention, the surgical tools are controlled by a surgeon with the aid of a specially designed robot \citep{Antico2019UltrasoundProcedures}, which requires a real-time understanding of the current task. Therefore, detecting the presence, location or pose of the surgical instruments may be useful in robotic surgeries as well \citep{Du2016CombinedSurgery}, \citep{Allan2013TowardSurgery}, \citep{Allan20183-DSurgery}. Finally, the actual location and movement of the tools can be extremely useful in rating the surgeries as well as generating an operative summary.

In order to track the surgical instruments, several approaches have been introduced, which use the signals collected during the procedure \citep{Elfring2010AssessmentSystems}, \citep{Reiter2012FeatureTools}. For instance, in vision-based methods, the instruments can be localized using the videos captured during the operation \citep{Wesierski2018InstrumentSurgeries}. These methods are generally reliable and inexpensive. Traditional vision-based methods rely on extracted features such as shape, color, the histogram of oriented gradients etc., along with a classification or regression method to estimate the presence, location or pose of the instrument in the captured images or videos \citep{Bouget2017Vision-basedLiterature}. However, these methods are dependent on pre-defined and painstakingly extracted hand-crafted features. Just logically defining and extracting such features alone is a major part of the detection process. Thus, these hand-crafted features and designs are not suitable for real-time applications.

Recent years have witnessed great advances in deep learning techniques in various computer vision areas such as image classification, object detection, and segmentation etc., and in medical imaging \citep{Litjens2017AAnalysisb}, due to the availability of large data and much improved computational power compared to the 1990s. The main advantage of deep learning methods over traditional computer vision techniques is that optimal high-level features can be directly and automatically extracted from the data. Therefore, there is a trend towards using these methods in analyzing the videos taken from laparoscopic operations \citep{Twinanda2017EndoNet:Videosb}.

Compared with the other surgical video tasks, detecting the presence and usage of surgical instruments in laparoscopic videos has certain challenges that need to be considered.

Firstly, since multiple instruments might be present at the same time, detecting the presence of these tools in a video frame is a multilabel (ML) classification problem. In general, ML classification is more challenging compared to the well-studied multiclass (MC) problem, where every instance is related to only one output. These challenges include but are not limited to using correlation and co-existence of different objects/concepts with each other and the background/context and the variations in the occurrence of different objects.

Secondly, as opposed to other surgical videos, such as cataract surgery \citep{AlHajj2019CATARACTS:Surgery}, robot-assisted surgery \citep{Sarikaya2017DetectionDetection} or videos from a simulation \citep{Zisimopoulos2017CanNetworks}, where the camera is stationary or moving smoothly, in laparoscopic videos, the camera is constantly shaking. Due to the rapid movement and changes in the field of view of the camera, most of the images suffer from motion blur and the objects can be seen in various sizes and locations. Also, the camera view might be blocked by the smoke caused by burning tissue during cutting or cauterizing to arrest bleeding. Therefore, using still images is not sufficient for detecting the instruments.

Thirdly, surgical operations follow a specific order of tasks. Although the usage of the tools doesn't strictly adhere to that order, it is nevertheless highly correlated with the task being performed. Using the information about the task and the relative position of the frame with regard to the entire video, the performance of the tool detection can be improved.

Lastly, since the performance of a deep classifier in a supervised learning method is highly dependent on the size and the quality of the labeled dataset, collecting and annotating a large dataset is a crucial task.

Endonet \citep{Twinanda2017EndoNet:Videosb} was the first deep learning model designed for detecting the presence of surgical instruments in laparoscopic videos, wherein Alexnet \citep{Krizhevsky2012ImageNetNetworksb} was used as a Convolutional Neural network (CNN), for feature extraction and is trained for the simultaneous detection of surgical phases and instruments. Inspired by this work, other researchers used different and more accurate CNN architectures with transfer learning \citep{Sahu2016ToolFeatures}, \citep{Prellberg2018Multi-labelNetworks} to classify the frames based on the visual features. For example, in \citep{Zia2016Fine-tuningDetection}, three CNN architectures are used, and \citep{Wang2017DeepVideos} proposed an ensemble of two deep CNNs.

\citep{Sahu2017AddressingCNN} were the first to address the imbalance in the classes in a ML classification of video frames. They balanced the training set according to the combinations of the instruments. The data were re-sampled to have a uniform distribution in label-set space and, class re-weighting was used to balance the data in a single class level. Despite the improvement gained by considering the co-occurrence in balancing the training set, the correlation of the tools’ usage was not considered directly in the classifier and the decision was made solely based on the presence of single tools.  \citep{AbdulbakiAlshirbaji2018SurgicalNetwork} used class weights and re-sampling together to deal with the imbalance issue.

In order to consider the temporal features of the videos, Twinanda et al. employed a hidden Markov model (HMM) in \citep{Twinanda2017EndoNet:Videosb} and Recurrent Neural Network (RNN) in \citep{Twinanda2017Vision-basedVideos}. Sahu et.al utilized a Gaussian distribution fitting method in \citep{Sahu2016ToolFeatures} and a temporal smoothing method using a moving average in \citep{Sahu2017AddressingCNN} to improve the classification results, after the CNN was trained.  \citep{Mishra2017LearningProcedures} were the first to apply a Long Short-Term Memory model (LSTM) \citep{Hochreiter1997LongMemoryb}, as an RNN to a short sequence of frames, to simultaneously extract both spatial and temporal features for detecting the presence of the tools by end-to-end training.

Other papers invoked different approaches to address the issues in detecting the presence of surgical tools. \citep{Hu2017AGNet:Detection} proposed an attention guided method using two deep CNNs to extract local and global spatial features. In \citep{AlHajj2018MonitoringNetworks}, a boosting mechanism was employed to combine different CNNs and RNNs. In \citep{Jin2018ToolNetworks}, the tools were localized by Faster RCNN \citep{Ren2015FasterNetworks} method, after labeling the dataset with bounding boxes containing the surgical tools.

It should be noted that none of the previous methods takes advantage of any knowledge regarding the order of the tasks and, the correlations of the tools are not directly utilized in identifying different surgical instruments.

In this paper, we propose a novel system called LapTool-Net to detect the presence of surgical instruments in laparoscopic videos. The main features of the proposed model are summarized as follows:
\begin{enumerate}
    \item Exploiting the spatial discriminating features and the temporal correlation among them by designing a deep Recurrent Convolutional Neural Network (RCNN)
    \item Taking advantage of the relationship among the usage of different tools by considering their co-occurrences
    \item The end-to-end training of the tool detector using a multitask learning approach
    \item Considering the inherent long-term pattern of the tools’ presence via a bi-directional RNN
    \item Using a small portion of the labeled samples considering the high correlation of the video frames to avoid overfitting
    \item Addressing the imbalance issue using re-sampling and re-weighting methods
    \item Providing state-of-the-art performance on a publicly available dataset on laparoscopic cholecystectomy
\end{enumerate}

The remainder of the paper is organized as follows: the main approach of LapTool-Net is described in section \ref{approach} and is elaborated in section \ref{methodolgy}. The performance of LapTool-Net is evaluated through experiments described in section \ref{results}. Section \ref{conclusion} concludes the paper.

\section{Approach}
\label{approach}
The uniqueness of our approach is based on the following three original ideas:
\begin{itemize}
    \item A novel ML classifier is proposed as a part of LapTool-Net, to take advantage of the co-occurrence of different tools in each frame – in other words, the context is taken into account in the detection process. In order to accomplish this objective, each combination of tools is considered as a separate class during training and testing and, is further used as a decision model for the ML classifier. To the best of our knowledge, this is the first attempt at directly using the information about the co-occurrence of surgical tools in laparoscopic videos in the classifier's decision-making.

    \item The ML classifier and the decision model are trained in an end-to-end fashion. For this purpose, the training is performed by jointly optimizing the loss functions for the ML classifier and the decision model using a multitask learning approach

    \item At the post-processing step, the trained model's prediction for each video is sent to another RNN to consider the order of the usage of different tools/tool combinations and long-term temporal dependencies – yet another consideration for the context.
\end{itemize}
 
 The overview of the proposed model is illustrated in Fig. 1. Let $\mathcal{D}=\{(x_{ij},Y_{ij})|0\leq i<m , 0<j<n\}$ be a ML dataset, where $x_{ij} \in \Re^d $ is the $i$th frame of the $j$th video and $Y_{ij}\subseteq \mathcal{Y}$ is the corresponding surgical instruments and $\mathcal{Y} \overset{\Delta}{=} \{y_1,y_2,...y_K\} $ is the set of all possible tools. Each subset of $\mathcal{Y}$ is called a label-set and each frame can have a different number of labels $|Y_{ij}|$. The tools associations can also be represented as a $K$ dimensional binary vector $y_{ij} = (y_1, y_2, ...,y_K) = \{0,1\}^K$, where each element is a 1 if the tool is present and a 0 otherwise. The goal is to design a classifier $F(x)$ that maps the frames of surgical videos, to the tools in the observed scene.
 
 \begin{figure*}
    \centering
    \includegraphics[width=\textwidth]{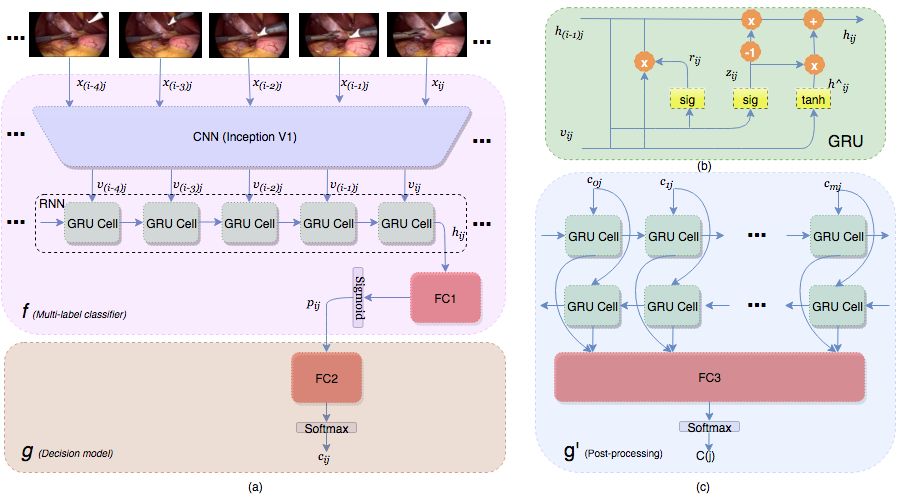}
    \caption{Block diagram of a) the proposed multiclass classifier F which consists of f and g, b) the architecture for Gated Recurrent Units (GRU) and c) The bi-directional RNN for post-processing. }
    \label{bd}
\end{figure*}

In order to take advantage of the combination of the surgical tools in a laparoscopic video, the well-known label power-set (LP) method is adopted in a novel way. The output of $F(x_{ij})$ is a label-set $\hat{Y}_{ij} \in \mathcal{\hat{Y}}$ (also called a superclass) of size $|\hat{Y}_{ij}| \leq K$, where $\mathcal{\hat{Y}}$ is the set of all possible subsets of $\mathcal{Y}$.

In order to calculate the confidence scores for each tool, along with the final decision, which is the class index in $\mathcal{\hat{Y}}$, the classifier $F$ is decomposed into $F(.)=g(f(.))$, where $g(f(x_{ij})):\Re^{K}\to\Re^{\hat{K}}$ is the decision model, which maps the confidence scores of the frame $i$ of the video $j$ to the label-set $\hat{Y}_{ij}$. The model $f$ takes the video frames as input and produces the confidence scores $P=(p_1,p_2,...p_K)=[0,1]^K$, where each element is the probability of the presence of one tool from the set $\mathcal{Y}$. It's worth mentioning that $f$ is as an ML classifier, while the output of the decision model $g$ is the label-set and therefore, classifier $F$ is an MC classifier based on the LPs.

The ML classifier $f$ consists of a CNN and an RNN. The CNN is responsible for extracting the visual features, while the RNN uses the sequence of features extracted by CNN and calculates the confidence scores $P$.

The output of the decision model $g$ for all the frames of each video forms a larger sequence $\Bar{C}$ of the model’s predictions. The sequence is used as the input to a bi-directional RNN $g'$ to exploit the long-term order of the tool usage.

The overall system is designed and tested using the dataset from M2CAI16\footnote{http://camma.u-strasbg.fr/m2cai2016/index.php/program-challenge} tool detection challenge1. The dataset contains 15 videos from cholecystectomy procedure, which is the surgery for removing the gallbladder. All the videos are labeled with 7 tools for every 25 frames. The tools are Bipolar, Clipper, Grasper, Hook, Irrigator, Scissors, and Specimen bags. There are 10 videos for training and 5 videos for validation.

The performance of LapTool-Net is measured through common metrics for ML and MC classification and a comparison is made with the current methods. The methodology derived from this approach is provided in more detail in the following section.

\section{Methodology}
\label{methodolgy}
\subsection{Multilabel Classification}
In ML classification, the goal is to assign multiple labels to each image. Higher dimensionality in label space and the correlation between the labels make the ML classification more challenging compared with MC problems. In the literature, two main approaches to deal with such issues in ML classification are accepted \citep{Gibaja2015ALearning}. One approach is called adaption, which aims at adapting existing machine learning models to deal with the requirements of ML classification. Since the output of an ML classifier is the confidence scores for each class, a decision policy is needed to make the final prediction. This decision is usually made based on top-k or thresholding methods.

The second paradigm for ML classification is based on problem transformation. The goal of problem transformation is to transform the ML problem into a more well-defined binary or MC classification. The most popular methods include binary relevance (BR), chain of classifiers \citep{Read2011ClassifierClassification} and LP. In BR, the problem is transformed into multiple binary classifiers for each class. This method doesn't take the dependencies of the classes into account. On the other hand, in a classifier chain, the binary classifiers are linked in a chain to deal with the classes’ correlations. In LP, multiple classes are combined into one superclass and the problem is transformed into an MC classification. The advantage of this method is that the class dependencies are automatically considered. However, as the number of classes increases, the complexity of the model increases exponentially. This is not an issue in laparoscopic videos. The reasons are 1) there is a limit for the number of tools in each frame (usually 3 or 4) and 2) the combinations of the tools are known. Since the LP method is more efficient than the classifier chain due to the use of just one classifier, it was determined to be more efficient for detecting the usage of surgical tools. Thus, we propose a novel classifier with LP being the decision layer for an ML classifier.

\subsection{Spatio-temporal Features}
\label{spatiotemporal}
In order to detect the presence of surgical instruments in laparoscopic videos, the visual features (intra-frame spatial and inter-frame temporal features) need to be extracted. We use CNN to extract spatial features. A CNN consists of an input layer, multiple convolutional layers, non-linear activation units, and pooling layers, followed by a fully connected (FC) layer to produce the outputs, which are typically the classification results or confidence scores. Each layer passes the results to the next layer and the weights for the convolutional and FC layers are trained using back-propagation to minimize a cost function. The output of the last convolutional layer is a lower dimensional representation of the input and therefore, can be considered as the spatial features. As shown in Fig \ref{bd}, the input frame $x_{ij}$ is sent through the trained CNN and the output of the last convolutional layer (after pooling) forms a fixed size spatial feature vector $v_{ij}$.

In the literature, several approaches have been proposed for utilizing the temporal features in videos for tasks such as activity recognition and object detection in videos \citep{Karpathy2014Large-scaleNetworks}, \citep{Simonyan2014Two-StreamVideos}. For instance, when there is a high correlation among video frames, it can be exploited to improve the performance of the tool detection algorithm.

An RNN is typically used to exploit the pattern of the instruments usage. It uses its internal memory (states) to process a sequence of inputs for time series and videos processing tasks \citep{Jin2018SV-RCNet:Network}. Although the motion features are not extracted explicitly when using the RNN, the temporal features are exploited through the correlation of spatial features in the neighboring frames.

Since the point of the RNN along with the CNN is to capture the high correlation among the neighboring frames, short sequences of frames (say 5 frames) are selected. Also, shorter sequences help the RNN have a better and faster convergence.

For each frame $x_{ij}$, the sequence of the spatial features $V_{ij}=[v_{(i-\lambda \Delta t)j} ...v_{(i-\Delta t)j} v_{ij}]$ is the input for the RNN, where the hyper-parameters $\lambda$ and $\Delta t$ are the number of frames in the sequence and the constant inter-frame interval respectively. The total length of the input is no longer than one second, which ensures that the tools remain visible during that time interval. Since the tool detection model is designed to be causal and to perform in real-time, only the previous frames with respect to the current frame can be used with the RNN.

We selected Gated Recurrent Unit (GRU) \citep{Cho2014LearningTranslation} as our RNN for its simplicity. The architecture is illustrated in Fig. \ref{bd}.(b) and formulated as:

\begin{align}
z_{ij}=&~\sigma (v_{ij}U^{z} + h_{i-\Delta t,j}W^{z}), \notag \\
r_{ij}=&~\sigma (v_{ij}U^{r} + h_{i-\Delta t,j}W^{r}),\notag \\ 
\tilde {h}_{ij}=&~tanh (v_{ij}U^{h}+ (r_{ij}\odot h_{i-\Delta t,j})W^{h}),\notag \\
h_{ij}=&~(1-z_{ij}) \odot h_{i-\Delta t,j} + z_{ij} \odot \tilde {h}_{ij} , 
\end{align}

where $U$ and $W$ are the GRU weights, $\odot$ is the element-wise multiplication and $\sigma$ is the sigmoid activation function. $z$ and $r$ are update gate and reset gate respectively. The final hidden state $h_{ij}$ is the output of the GRU and is the input to a fully connected neural network $FC1$. The output layer $FC1$ is of size $K$ (the number of tools) and after applying the sigmoid function, produces the vector of confidence scores $P(ij)$ for all classes.

We designed the above RCNN architecture as the ML classifier model f shown in Fig. \ref{bd}.a, which exploits the spatiotemporal features of a given input frame and produces the vector of confidence scores of all the tools, which in turn is the input to the decision model $g$.

\subsection{Decision Model}
\label{decision}
One of the main challenges in ML classification is effectively utilizing the correlation among different classes. Using LP (as described earlier), uncommon combinations of the classes will automatically be eliminated from the output and the classifier's attention is directed towards the more possible combinations.

\begin{figure}[t]
\centering
{\includegraphics[width=3.5in]{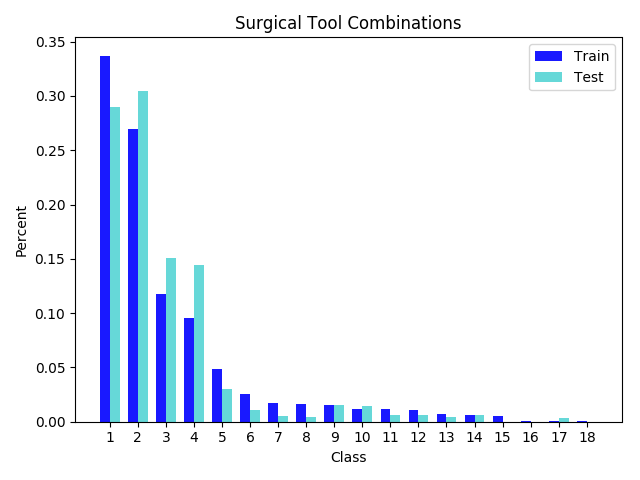}}%
\caption{The distribution for the combination of the tools in M2CAI dataset}
\label{combination}
\end{figure}

As mentioned before, not all the $2^K$ combinations are possible in a laparoscopic surgery. Fig. \ref{combination} shows the percentage of the most likely combinations in the M2CAI dataset. The first 15 classes out of a possible maximum of 128 span more than 99.95\% of the frames in both the training and the validation sets, and the tools combinations have almost the same distribution in both cases.

Since an LP classifier is MC, the cost function for training a machine learning algorithm has to be the conventional one-vs-all (categorical) loss. For example, Softmax cross-entropy (CE) is the most popular MC loss function. However, Softmax CE requires the classes to be mutually exclusive, which is not true in the LP method. In other words, while using a Softmax loss, each superclass is treated as a separate class, i.e. separate features activate a superclass. This causes performance degradation in the classifier and therefore, more data is required for training. We address this issue by a novel use of LP as the decision model $g$, which we apply to the ML classifier $f$. Our method helps the classifier to consider our superclasses as the combinations of classes rather than separate mutually exclusive classes.

The decision model is a fully connected neural network ($FC2$), which takes the confidence scores of $f$ and maps them to the corresponding superclass. When the Softmax function is applied, the output of $g(.)$ is the probability of each superclass $Q=(q_1,q_2,...,q_{\hat{K}})$ where $\hat{K}$ is the size of the superclass set. The final prediction of the tool detector $F$ is the index of the superclass with the highest probability and for frame $i$ of video $j$ is calculated as:

\begin{equation}
    {c}_{ij} = argmax(Q_{ij})
    \label{final-pred}
\end{equation}

\subsection{Class Imbalance}

Class imbalance has been a well-studied area in machine learning for a long time \citep{Buda2018ANetworks}. It is known that in skewed datasets, the classifier's decision is inclined towards the majority classes. Therefore, it is always beneficial to have a uniform distribution for the classes during training. Two major approaches have been proposed in the literature to deal with imbalanced datasets.

One approach is called cost sensitive and is mainly based on class re-weighting. In this method, the outputs of the classes or the loss function during training are weighted based on the frequency of the classes. Although this approach works in some cases, the choice of the weights might not depend solely on the distribution of the data, since the complexity of the input is not known before training. Thus, class weights are another set of hyper-parameters that needs to be determined.

Another solution to an imbalanced dataset is to change the distribution in the input. This can be accomplished using over-sampling for the minority classes and under-sampling for the majority classes. However, in ML classification, finding a balancing criterion for re-sampling is challenging \citep{Charte2015AddressingAlgorithms}, since a change in the number of samples for one class might affect other classes as well.

The number of samples for each tool before balancing is shown in Table \ref{imbalance}. In order to overcome this issue, we perform under-sampling to have a uniform distribution of the combination of the classes. The main advantage of under-sampling over other re-sampling methods is that it can also be applied to avoid overfitting caused by the high correlation between the neighboring frames of a laparoscopic video. Therefore, we try different under-sampling rate to find the smallest training set without sacrificing the performance.

\begin{table}[ht]
    \caption{Number of frames for each tool in M2CAI}
    \centering
    \begin{tabular}{c|c|c}
        Tool & Train & Validation \\
        \hline
         Bipolar & 631 & 331\\
         Clipper & 878 & 315 \\
         Grasper & 10367 & 6571\\
         Hook & 14130 & 7454\\
         Irrigator & 953 & 131\\
         Scissors & 411 & 158\\
         Specimen Bag & 1504 & 483\\
         no tools & 2759 & 1888 \\
         \hline
         \hline
         total & 23421 & 12512\\
    \end{tabular}
    
    \label{imbalance}
\end{table}

Since this approach will not guarantee balance, a cost-sensitive weighting approach can be used along with an ML loss, prior to the LP decision layer; nonetheless, we empirically found that this doesn't affect the performance of the ML classifier.

\begin{figure*}[t]
    \centering
    \subfloat[before balancing]{\includegraphics[width = 0.44\textwidth]{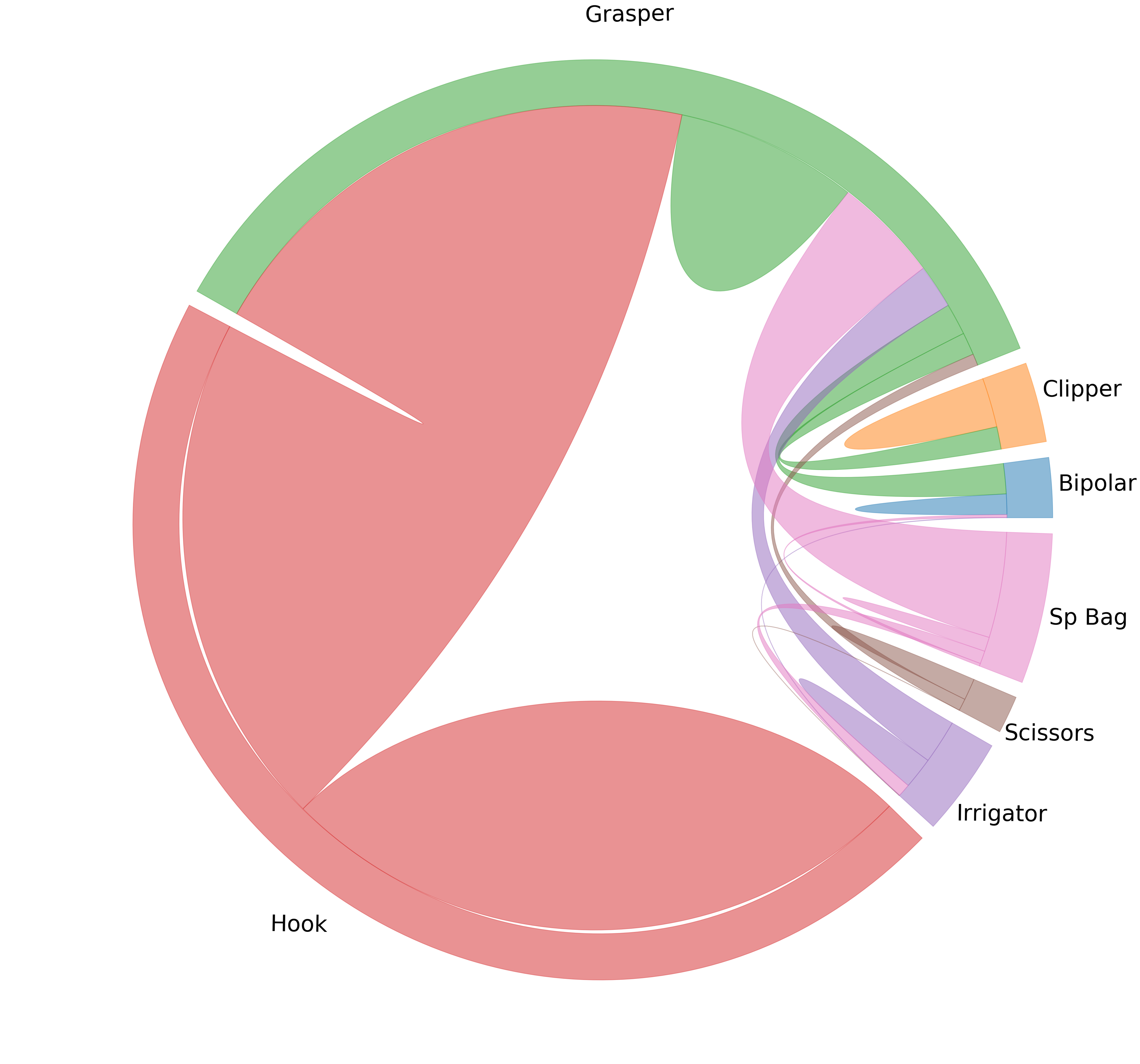}} 
    \subfloat[after balancing]{\includegraphics[width = 0.44\textwidth]{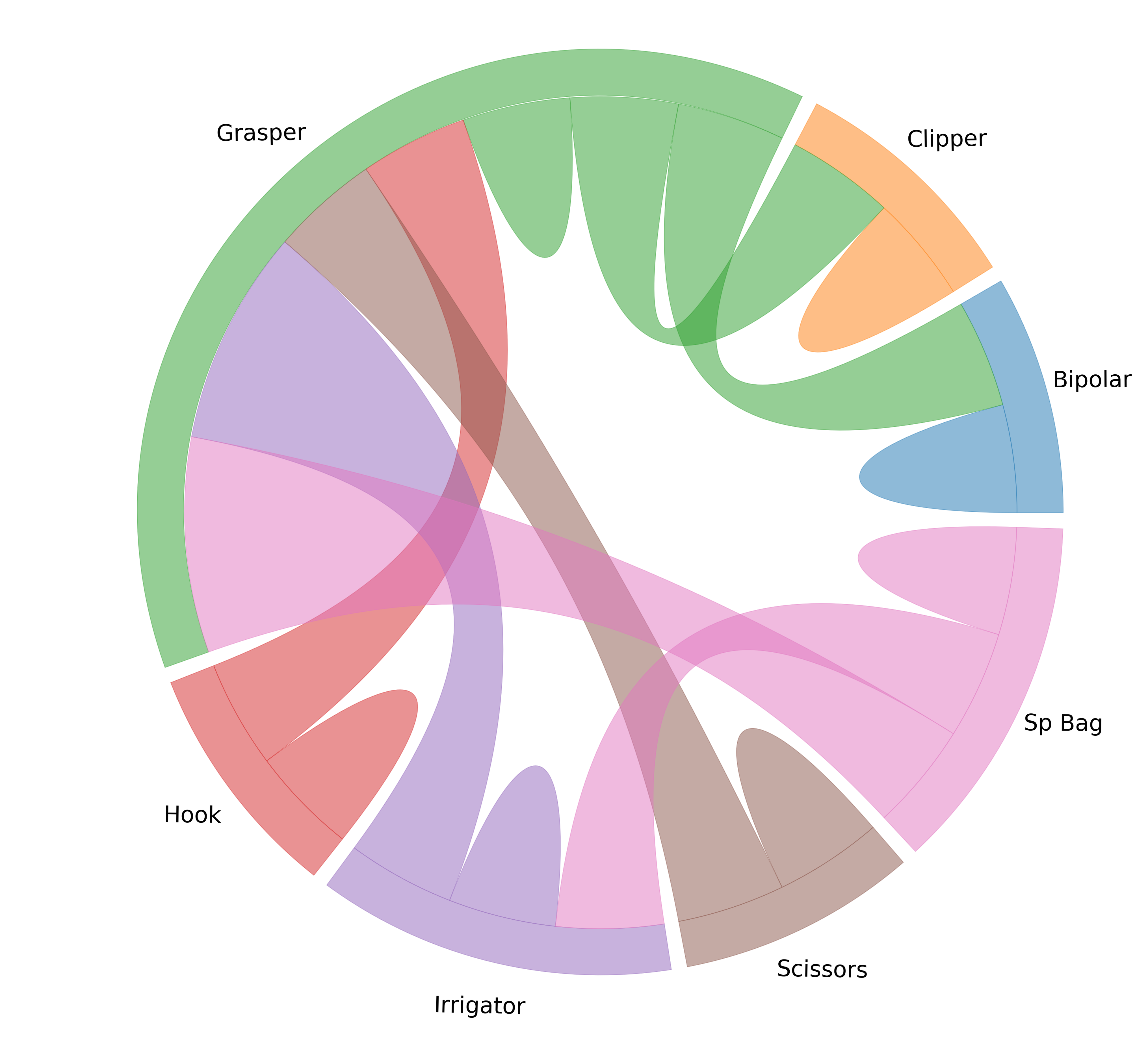}}\\
    \caption{The chord diagram for the relationship between the tools before and after balancing based on the tools' co-occurrences}
    \label{chord diagram}
\end{figure*}

Figure \ref{chord diagram} shows the relationship among the tools after re-sampling. It can be seen that the LP-based balancing method not only tends to a uniform distribution in the superclass space, it also improves the balance of the dataset in the single class space (with the exception of Grasper, which can be used with all the tools).
\subsection{Training}

Since our tool detector $F(x)$ is decomposed into an ML classifier $f$ and an MC decision model $g$, the requirement of both models needs to be considered during training. In order to accomplish this, the model is trained using both ML and MC loss functions.

We propose to use joint training paradigm for optimizing the ML and MC losses as a multitask learning approach. In order to do that, two optimizers are defined based on the two losses with separate hyper-parameters such as learning rate and trainable weights. Using this technique, the extraction of the features is accomplished based on the final prediction of the model.

Having the vector of the confidence scores $P$, the ML loss $L_f$ is the sigmoid cross-entropy and is formulated as:

\begin{equation}
    L_{f} = - \frac{1}{d}\sum_{x\in \mathcal{D}}\log (p_{k=Y}) ,          \quad \quad P=(\sigma f(x))
    \label{ML loss}
\end{equation}
where $Y$ is the correct label for frame $x$, $d$ is the total number of frames and $\mathcal{D}$ is the training set.

The Softmax CE loss function $L_g$ for the decision model is formulated as:

\begin{equation}
    L_{g} = - \frac{1}{d}\sum_{x\in \mathcal{D}}\log (q_{k=\hat{Y}}), \quad \quad Q=(softmax(g(f(x))))
    \label{MC loss}
\end{equation}

The total loss function is the sum of the two losses and formulated as:

\begin{equation}
    \mathcal{L} = L_{f}+\beta L_{g}
    \label{total loss}
\end{equation}
where $\beta$ is a constant weight for adjusting the impact of the two loss functions. The training is performed in an end-to-end fashion using the backpropagation through time method.

\subsection{Post-processing}

The final decision of the RCNN model from the previous section is made based on the extracted spatio-temporal features from a short sequence of frames. In other words, the model benefits from a short-term memory using the correlation among the neighboring frames. However, due to the high under-sampling rate for the balanced training set, this method might not produce a smooth prediction over the entire duration of the laparoscopic videos. In order to deal with this issue, we model the order in the usage of the tools with an RNN over all the frames of each video \citep{Namazi2018AutomaticVideos}.

Due to memory constraints, the final prediction from equation \ref{final-pred} of the RCNN, $\bar {C}(j) = [c_{0j} ....c_{mj} ]$] for all the videos $0<j<n$, is selected as the input for the post-processing RNN. Since not all the videos have the same length, the shorter videos are padded with the no-tool class.

Our post-processing occurs offline, after the surgery is finished. Therefore, future frames can also be used along with past frames to improve the classification results of the current frame. In order to accomplish this, a bi-directional RNN is employed, which consists of two RNNs for the forward and backward sequences. The backward sequence is simply the reverse of $\bar {C}$ . The outputs of the bi-RNN are concatenated and fed to $FC3$ for the final prediction ($g\prime$ in Fig. \ref{bd}.c).

Since the input frames for the bi-RNN are in a specific order, it's not possible to balance the input through re-sampling. Therefore, class re-weighting is performed to compensate for the minority classes. The class weights are chosen to be proportional to the inverse of the frequency of the superclasses. The loss function is:
\begin{equation}
    L_p = - \frac{1}{d}\sum_{x\in \mathcal{D}} (w_{k=\hat{Y}}) \log \bar q_{k=\hat{Y}}, \quad \bar{Q}=(softmax(g \prime (\bar {C})))
\end{equation}
where $w_k$ is the weight for the superclass $k$, $g\prime$ is the bi-RNN with 64 hidden layers and $\bar{Q}=(\bar{q}_{1}, ..., \bar{q}_{\hat{K}})$ is the superclass probability vector.

\section{Results}
\label{results}
In this section, the performance of the different parts of the proposed tool detection model on M2CAI dataset is validated through numerous experiments using the appropriate metrics.

We selected Tensorflow \citep{AbadiTensorFlow:Systemsb} for all of the experiments. The CNN in all the experiments was Inception V1 \citep{Szegedy2015GoingConvolutions}. In order to have better generalization, extensive data augmentation, such as random cropping, horizontal and vertical flipping, rotation and a random change in brightness, contrast, saturation, and hue were performed during training. The initial learning rate was 0.001 with a decay rate of 0.7 after 5 epochs and the results were taken after 100 epochs. The batch size was 32 for training the CNN models and, 40 for the RNN-based models. All the experiments were conducted using an Nvidia TITAN XP GPU. The source code of the project is available on Github.

\subsection{Metrics}

Since the proposed model is MC, the corresponding evaluation metrics were chosen. Due to the high imbalance of the validation dataset, accuracy alone is not sufficient to evaluate the proposed model. Therefore, we used F1 score to compare the performance of different models in both per-class and overall metrics. These are calculated as:

\begin{equation}
    F1_{micro} = 2\frac{P_{pc}.R_{pc}}{P_{pc}+R_{pc}}, \\
    F1_{macro} = 2\frac{P_{ov}.R_{ov}}{P_{ov}+R_{ov}}
\end{equation}
where $P_{pc}$, $R_{pc}$, $P_{ov}$ and $R_{ov}$ are per-class precision/recall and overall precision/recall respectively and are calculated as:

\begin{equation}
     P_{pc} = \frac{1}{K}\sum_{k=1}^{K}\frac{N_{y_{k}}^{c}}{N_{y_{k}}^{p}},
     R_{pc} = \frac{1}{K}\sum_{k=1}^{K}\frac{N_{y_{k}}^{c}}{N_{y_{k}}}\\
 \end{equation}
 \begin{equation}
     P_{ov} = \frac{\sum_{k=1}^{K}{N_{y_{k}}^{c}}}{\sum_{k=1}^{K}{N_{y_{k}}^{p}}},     R_{ov} = \frac{\sum_{k=1}^{K}{N_{y_{k}}^{c}}}{\sum_{k=1}^{K}{N_{y_{k}}}}\\
 \end{equation}
where $N_{y_{k}}^{c}$ , $N_{y_{k}}^{p}$ and $N_{y_{k}}$ are the number of correctly predicted frames for class $k$, the total number of frames predicted as $k$, and the total number frames for class $k$. Only frames with all the tools predicted correctly are considered exact matches.

In order to evaluate the RCNN model $f$, we used ML metric - mean Average Precision (mAP), which is the mean of average precision (a weighted average of the precision with the recall at different thresholds) for all 7 tools.

\subsection{CNN Results}

In the first experiments, we assumed that the classifier $f$ is a CNN and the decision model $g$ is applied to the resulting scores from the CNN. Since the dataset was labeled only for one frame per second (out of 25 frames/sec), there was a possibility of using the unlabeled frames for training, as long as the tools remain the same between two consecutive labeled frames. We used this unlabeled data to balance the training set, according to the LPs. The CNN was trained with the loss function \ref{ML loss} with $FC1$ of size 7. Table \ref{CNN-balancing} shows the results of our CNN with different training set sizes for the tools listed in Table \ref{imbalance}.

\begin{table}[t]
    \centering
    \caption{Results for the multi-label classification of the CNN}
    \begin{tabular}{c|c|c|c|c}
        Total Frames & Balanced & Acc(\%) & mAP(\%) & F1-macro(\%) \\
         \hline
         23k & No & 77.23 & 61.02 & 58.48 \\
         150k & Yes & 75.90 & 71.15 & 70.49 \\
         75k & Yes & 74.78 & 77.24 & 74.81 \\
         25k & Yes & 75.40 & 78.58 & 74.64 \\
         6k & Yes & 74.36 & 78.22 & 74.43 \\%
         3k & Yes & 73.10 & 73.69 & 70.85 \\
         
    \end{tabular}
    \label{CNN-balancing}
\end{table}

As was to be expected, the unbalanced training set results shown in the first row of Table \ref{CNN-balancing} has the lowest performance on all the metrics. The high exact match accuracy (Acc) of 77.23\% and the lower results on per-class metrics, such as F1-macro and mAP show that the model correctly predicted the majority classes (grasper and hook) but has poor performance for the less used tools such as scissors.

In order to balance the datasets, the following specific steps were taken: 1) 15 superclasses were selected and the original frames were re-sampled to have a uniform distribution in the set of label-sets $\mathcal{\hat{Y}}$. The numbers of frames for each superclass were randomly selected to be 10,000, 5,000, 1,666, 400 and 200. 2) Multiple copies of some frames were copied and pasted to the final set in the first two training sets, because of the availability of fewer frames in some tools such as scissors. This accomplished the intended over-sampling. 3) Similarly, under-sampling was performed in at least one class in all sets and, in all classes in the last two sets, because too many frames were available for some tools.

Under these conditions, we can discuss the results presented in rows 2 through 6 in Table \ref{CNN-balancing}. While the exact match accuracy is the highest in the 150K set, it has the lowest score on the per-class metrics. The likely reason is the high over-sampling rate, which causes overfitting for the less frequently occurring classes. On the other hand, a very high under-sampling rate in the 3K set results in lower accuracy, likely due to the lack of informative samples.

The best per-class results are for the 25K/6K versus the 150K/75K, which is due to the lower correlation among the inputs of the CNN during training. We used the 6K dataset for the rest of the experiments versus the 25K, because adding the RNN and decision model to the selected CNN would increase the size of the model (RCNN-LP), and the chances of overfitting increases.

In order to evaluate the effect of utilizing the co-occurrence of different surgical tools, we tested the LP method as the primary classifier, as well as the decision model, using different training strategies. The configurations for each experiment are shown in Table \ref{configs}.

\begin{table*}[ht]
    \centering
    \caption{Setup configurations for training the multiclass CNN}
    \begin{tabular}{c|c c c c c}
        
        Exp. num  & Loss function & FC1 size & FC2 size & Trainable weights & Training method\\
        \hline
        1 & (\ref{MC loss}) & 15 & - & all & - \\
        2 (150k) & (\ref{MC loss}) & 15 & - & all & - \\
        3 & (\ref{ML loss})/(\ref{MC loss}) & 15 & 15 & CNN+FC1/ FC2 & Sequential\\
        4 & (\ref{ML loss})/(\ref{MC loss}) & 7 & 15 & CNN+FC1/ CNN+FC2 & Alternate\\
        5 & (\ref{ML loss})/(\ref{MC loss}) & 7 & 15 & CNN+FC1/ all & Alternate\\
        6 & (\ref{ML loss})/(\ref{MC loss}) & 7 & 15 & CNN+FC1/ FC2 & Sequential \\
        7 & (\ref{total loss}) & 7 & 15 & all & Joint 
    \end{tabular}
    \label{configs}
\end{table*}

In sequential training, the CNN was trained first, and the decision model was added on top of the trained model, while the CNN weights remained unchanged. In alternate training, the trainable weights change with the loss at every other step. The joint training method is explained in the previous section. We used MC metrics; exact match accuracy, micro and macro F1, and average per-class precision and recall. The results are shown in Table \ref{exp0} and the precision and recall for each tool are shown in Table \ref{PR}.

\begin{table}[ht]
    \centering
    \caption{Results for the multiclass CNNs}
    \begin{tabular}{c|c c c c c}
        
        Exp. num & Acc(\%) & F1-macro(\%) & F1-micro(\%) & Mean P(\%) & Mean R(\%)\\
        \hline
        1 & 70.01 & 69.14 & 84.57 &72.90 &67.98 \\
        2 & 76.13 & 73.77 & 87.89 & 86.08 & 67.24\\
        3 & 73.18 & 74.30 & 86.92 & 79.65 & 70.80\\
        4 & 74.42 & 75.70 & 87.75 & 82.37 & 71.48\\
        5 & 72.44 & 75.23 & 86.42 & 87.60 & 67.25\\
        6 & 74.97 & 75.47 & 88.04 & 80.67 & 73.21\\
        7 & 76.31 & 78.32 & 88.53 & 78.48 & 78.95\\
    \end{tabular}
    \label{exp0}
\end{table}

\begin{table*}[t]
    \centering
    \caption{Precision (P(\%)) and Recall (R(\%)) of each tool for the multiclass CNNs}
    \begin{tabular}{|c|c|c|c|c|c|c|c|c|}
        \hline
          & \multicolumn{2}{c|}{Bipolar} & \multicolumn{2}{c|}{Clipper} & \multicolumn{2}{c|}{Grasper} & \multicolumn{2}{c|}{Hook}     \\
        Exp. & P & R & P & R & P & R & P & R  \\
        \hline
        1 &71.2&66.0&72.7&58.4&90.3&70.9&92.8&90.6\\
        \hline
        2 &76.5&35.7&85.4&52.0&92.0&80.3&95.2&90.0\\
        \hline
        3 &84.4&71.5&78.0&57.4&91.0&75.8&94.6&90.9\\
        \hline
        4 &81.5&70.8&80.7&60.0&89.9&79.8&95.3&90.1\\
        \hline
        5 &91.7&56.5&81.2&53.6&91.3&75.9&97.5&86.1\\
        \hline
        6 &75.1&75.1&69.2&59.3&90.0&81.5&95.2&90.3\\
        \hline
        7 & 83.9&72.6&72.6&74.2&91.1&81.2&94.3&91.4\\
        \hline
    \end{tabular}
    \label{PR}
\end{table*}

\begin{table*}[t]
    \centering
    \begin{tabular}{|c|c|c|c|c|c|c|}
        \hline
          & \multicolumn{2}{c|}{Irrigator} & \multicolumn{2}{c|}{Scissors} & \multicolumn{2}{c|}{Specimen bag}   \\
        Exp. & P & R & P & R & P & R  \\
        \hline
        1 &56.7&65.5&61.9&32.9&64.6&91.4\\
        \hline
        2 &85.8&74.6&93.8&48.1&73.6&89.8\\
        \hline
        3 &57.0&53.2&79.6&54.4&72.8&92.1\\
        \hline
        4 &63.3&56.5&88.7&50.0&77.0&93.0\\
        \hline
        5 &78.2&59.0&93.1&51.8&79.9&87.5\\
        \hline
        6 &72.8&70.4&87.8&41.1&74.3&94.4\\
        \hline
        7 &59.7&75.4&71.6&63.9&75.8&93.7\\
        \hline
    \end{tabular}
\end{table*}

In the first two experiments, the LP method was used directly to map the video frames to the corresponding superclass. In order to accomplish this, the features extracted using CNN were connected to an FC layer of size 15 and the network was trained with the loss function from equation \ref{MC loss}. We selected the balanced training sets from the previous experiments with 6K and 150K samples. It can be seen from experiment 1 and 2 (Table \ref{exp0}), which correspond to 6K sample and 150k samples respectively, that both accuracy and F1 scores increase, when the training set is larger. Also, the precision and recall in Table \ref{PR} show some improvements in almost all classes. However, compared with the results from Table \ref{CNN-balancing}, we observe minor improvements in accuracy and F1, when using 150k frames with LP classifier, while the metrics decrease with a smaller training set. Considering both training sets were balanced based on LP, the observation suggests that the LP-based classifier needs more examples for reasonable performance. This is because, in an LP classifier, the superclasses are treated as separate classes with different features from the corresponding single label classes, which requires the classifier to have more training examples to learn the discriminating features. This can also be confirmed by checking the relatively close precision/recall for grasper and hook in Table \ref{PR}, which have more unique frames (due to lower under-sampling rate), in the two experiments.

In experiment 3, ML loss was tried instead of MC for training the LP classifier with 15 superclasses. $FC2$ was added as a decision model and was trained sequentially. As shown in Table \ref{exp0}, the per-class F1 score for experiment 3 improves compared to experiment 1, while the exact match accuracy is lower. This is probably because the model is still not aware that a superclass is a combination of multiple classes.

In experiments 4, 5 and 6, the CNN was trained using ML loss \ref{ML loss} with 7 classes and the decision layer was added on top of the confidence scores. We evaluated the model using different training strategies. All three of these experiments produced better results than experiment 3. This is likely because the model can learn the pattern of the 7 tools easier with the ML loss, compared with learning the pattern for the combination classes using 15 classes.

The point of performing the experiments 4 and 5 was to evaluate the effect of the decision model in training the feature extractor and ML classifier. In both experiments, the decision model $g$ and the CNN were trained alternately. The weights of the CNN were frozen in experiment 6, while in experiments 4 and 5 they were trained at each step. Therefore, in experiment 6, the role of the decision model was to just use the co-occurrence information to find the correct classes (superclasses) using the confidence scores of a trained model. The results show improvement in F1 scores in all three experiments compared with the results from Table \ref{CNN-balancing}. This is due to the fact that using LP as the decision model, the co-occurrence of surgical tools in each frame is considered directly in the classification method without learning separate patterns for superclasses.

In experiment 7, the loss is the weighted sum of the ML and MC loss functions and the training was performed on all the weights of the model. We can see that the end-to-end training of the RCNN-LP produces significantly better results compared with all other training methods, such as sequential and alternate training. The reason is that in end-to-end training, all parts of the model is trained simultaneously to reach better confidence scores and hence, better final decision.

\subsection{LapTool-Net Results}

In this section, the performance of the proposed model is evaluated after considering the spatio-temporal features using an RNN. Similar to the previous section, we tested the model before and after adding the decision model. The dataset for training is the 6K balanced set and all the models were trained end-to-end. For training the RCNN model, we used 5 frames at a time (current frame and 4 previous frames) with an inter-frame interval of 5, which resulted in a total distance of 20 frames between the first and the last frame. The RCNN model was trained with a Stochastic Gradient Descent (SGD) optimizer. The data augmentation for the post-processing model includes adding random noise to the input and randomly dropping frames to change the duration of the sequences.

Table \ref{rnn0} shows the results of the proposed LapTool-Net. For ease of comparison, we have copied the results from the previous section for the CNN with and without the LP decision model. It can be seen that by considering the temporal features through the RCNN model the exact match accuracy and F1 scores were improved. The higher performance of the LapTool-Net is due to the utilization of the frames from both the past and the future of the current frame, as well as the long-term order of the usage of the tools, by the bi-directional RNN.

\begin{table}[ht]
    \centering
    \caption{Final results for the proposed model}
    \begin{tabular}{c|c|c|c}
        & Acc(\%) & F1-macro(\%) & F1-micro(\%)\\
        \hline
        CNN & 74.36 & 74.43 & 87.70\\
        
        CNN-LP & 76.31 & 78.32 & 88.53\\
        
        RCNN & 77.51 & 81.95 & 89.54\\
        
        RCNN-LP & 78.58 & 84.89 & 89.79\\
        
        LapTool-Net & 80.96 & 89.11 & 91.35\\
    \end{tabular}
        \label{rnn0}
\end{table}
The precision, recall and F1 score for each of the tools are shown in Table \ref{final-pred}. Compared with the results from Table \ref{PR}, we can see that the F1 score for clippers and scissors have significantly increased, because there is a high correlation between the usage of these tools and the tasks, i.e. the order in the occurrence of the tools (e.g. cutting only happens after clipping is completed). The lowest performance is for Irrigator, which is probably because of the irregular pattern in its use (only used for bleeding and coagulation, which can occur any time during the surgery). The higher over-all recall is likely because of the class re-weighting method. We believe the performance could improve with a better choice of the weights.

\begin{figure*}[t]
    \centering
    \subfloat[Grasper]{\includegraphics[width = 0.95in]{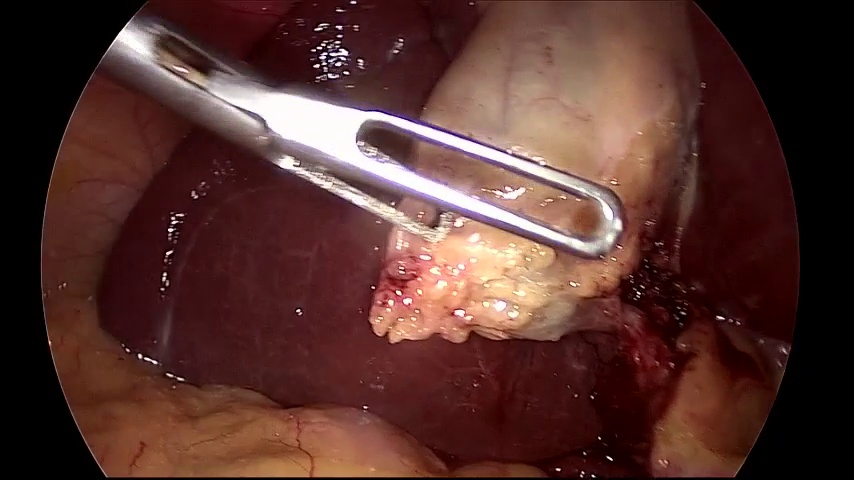}}
    \subfloat[Grasper/Hook]{\includegraphics[width = 0.95in]{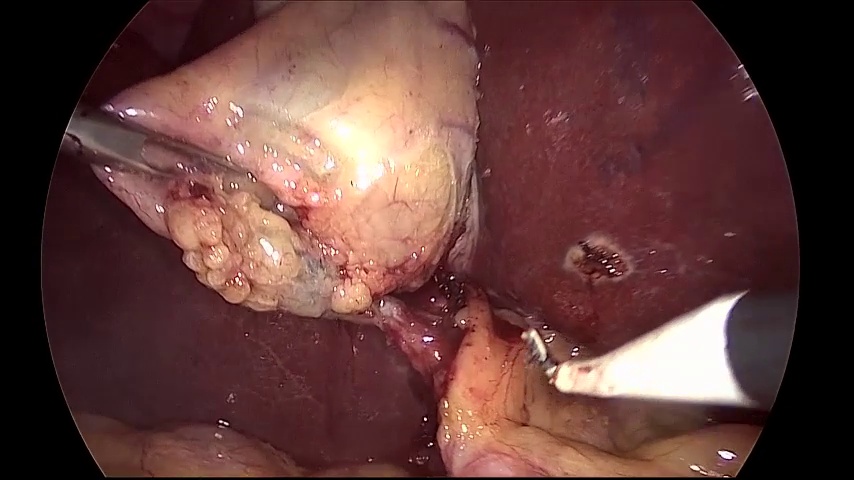}}
    \subfloat[Grasper/Clipper]{\includegraphics[width = 0.95in]{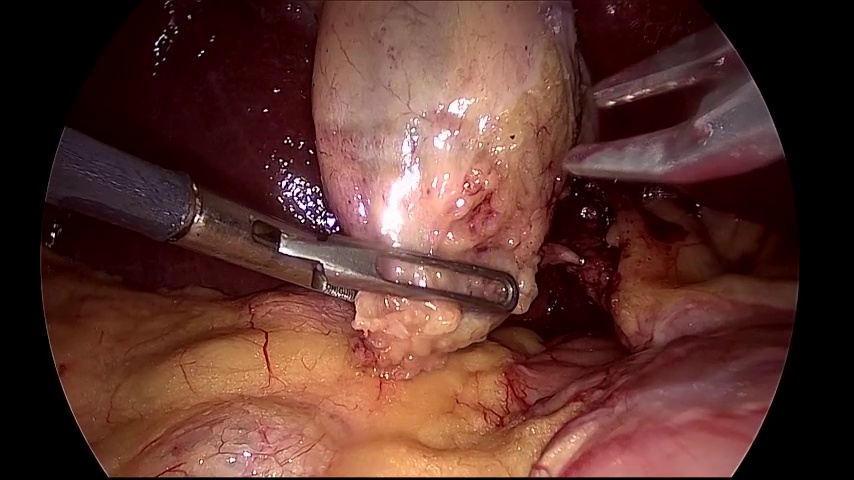}}
    \subfloat[Grasper/Scissor]{\includegraphics[width = 0.95in]{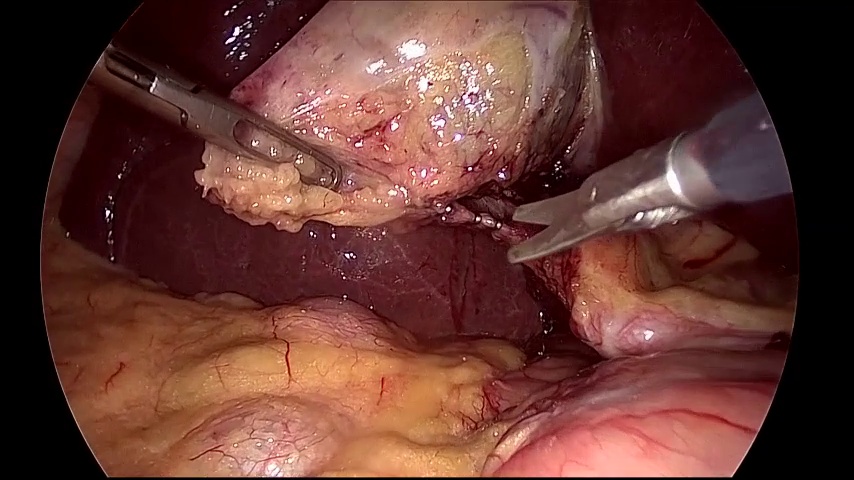}}
    \subfloat[Grasper/Hook]{\includegraphics[width = 0.95in]{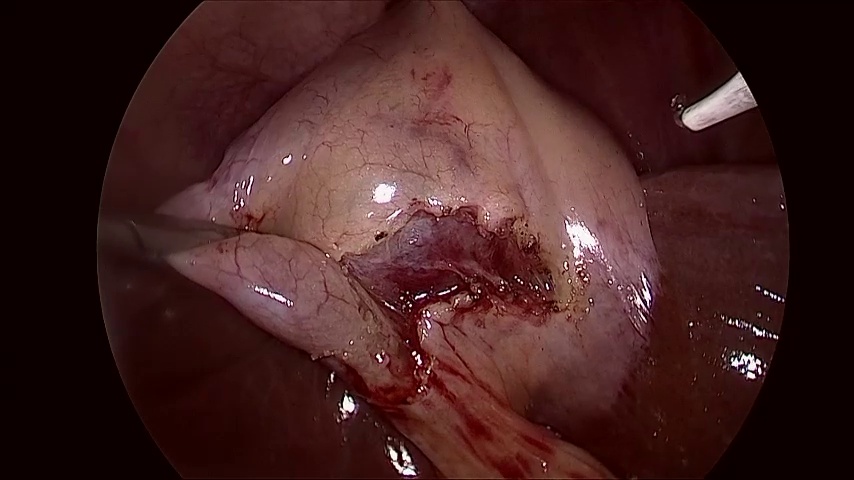}}
    \subfloat[Grasper/Bipolar]{\includegraphics[width = 0.95in]{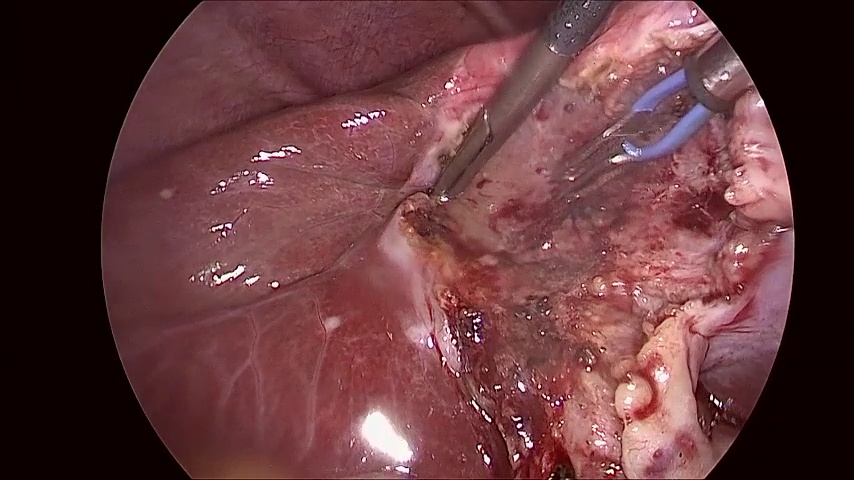}}\\
    
    \subfloat[Grasper]{\includegraphics[width = 0.95in]{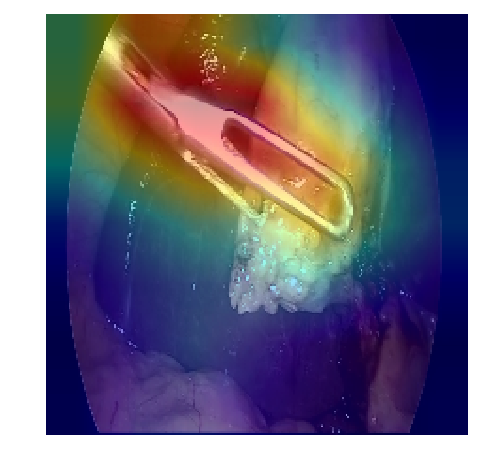}}
    \subfloat[Hook]{\includegraphics[width = 0.95in]{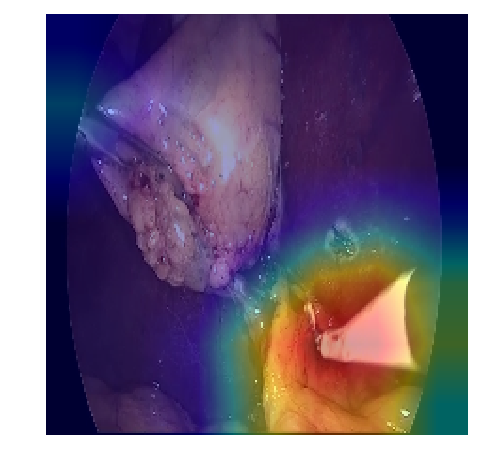}}
    \subfloat[Grasper]{\includegraphics[width = 0.95in]{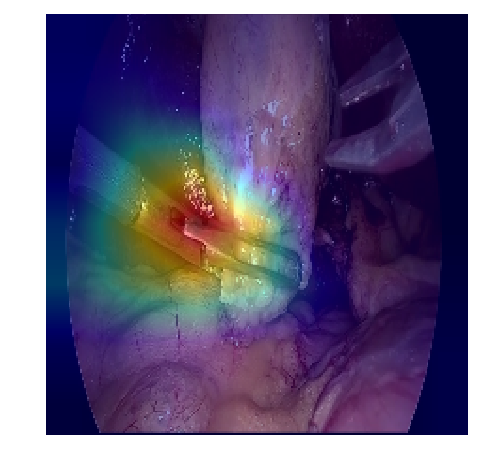}}
    \subfloat[Scissors]{\includegraphics[width = 0.95in]{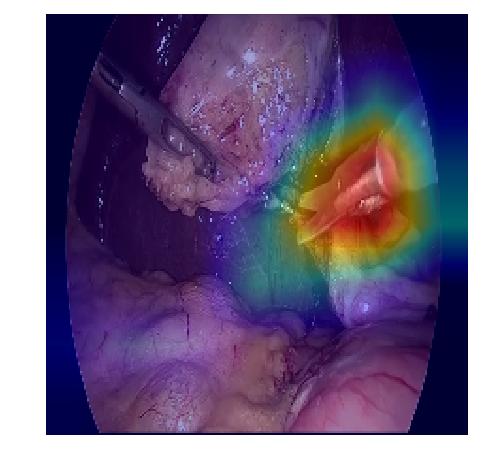}}
    \subfloat[Hook]{\includegraphics[width = 0.95in]{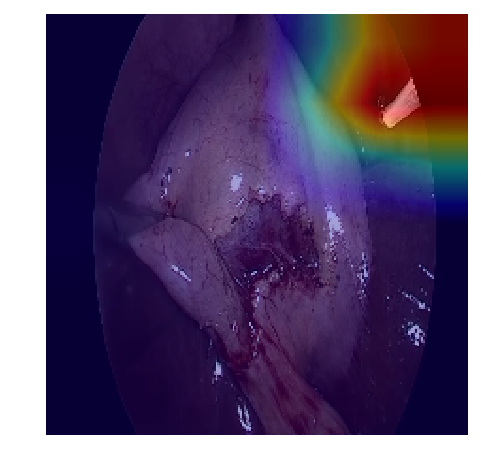}}
    \subfloat[Bipolar]{\includegraphics[width = 0.95in]{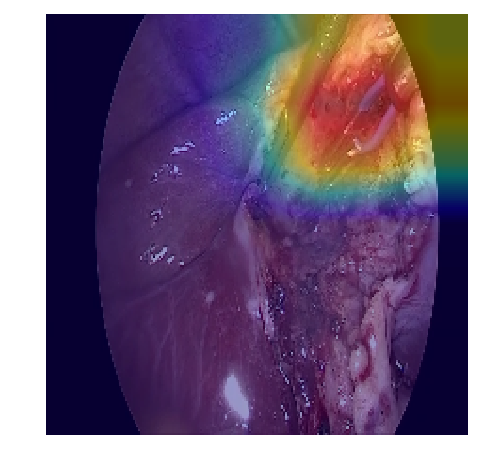}}\\
  \caption{The visualization of the class activation maps for some examples, based on the prediction of the model}
\label{visualization}  
\end{figure*}

\begin{table}[htbp]
    \centering
    \caption{The precision, recall and F1 score of each tool for LapTool-Net}
    \begin{tabular}{c|c|c|c}
        Tool & Precision(\%) & Recall(\%) & F1(\%)\\
        \hline
        Bipolar & 0.82 & 0.95 & 0.88\\
        
        Clipper & 0.85 & 0.98 & 0.91\\
        
        Grasper & 0.89 & 0.88 & 0.89\\
        
        Hook & 0.94 & 0.94 & 0.94\\
        
        Irrigator & 0.74 & 0.91 & 0.82\\
        
        Scissors & 0.82 & 0.99 & 0.90\\
        
        Specimen bag & 0.92 & 0.89 & 0.91\\
        \hline
        \hline
        Mean & 0.85 & 0.93 & 0.89\\
    \end{tabular}
    \label{final-results}
\end{table}
In order to localize the predicted tools, the attention maps were visualized using grad-CAM method \citep{Selvaraju2017Grad-CAM:Localization}. The results for some of the frames are shown in figure \ref{visualization}. In order to avoid confusion with frames that multiple tools, only the class activation map of a single tool is shown based on the prediction of the model. The results show that the visualization of the attention of the proposed model can also be used in reliably identifying the location of each tool.

\subsection{Comparison with Current Work}

In order to validate the proposed model, we compared it with previously published research on the M2CAI dataset. The result is shown in Table \ref{comparison}. Since all the methods reported their results using ML metrics such as mAP, we compared our ML classifier $f$, which is the RCNN model, along with the final model. We show that our model out-performed previous methods by a significant margin even when choosing a relatively shallower model (Inception V1) and while using less than 25\% of the labeled images.

\begin{table*}[t]
    \centering
    \caption{Comparison of tool presence detection methods on M2CAI}
    \begin{tabular}{c|c|c|c}
    
        Method & CNN & mAP(\%) & F1-Macro(\%)\\
    \hline    
        \bf{LapTool-Net} & \bf{Inception-V1} & - & \bf{89.11}\\
        
        \bf{RCNN (ours)} & \bf{Inception-V1} & \bf{87.88} & \bf{81.95}\\
        
        \citep{Hu2017AGNet:Detection} & Resnet-101 \citep{He2016DeepRecognitionb} & 86.9 & -\\
        
        \citep{Sahu2017AddressingCNN} & Alexnet & 65 & -\\
        
        \citep{Wang2017DeepVideos} & Inception-V3 \citep{Szegedy2016RethinkingVision} & 63.8 & -\\
        
        \citep{Sahu2016ToolFeatures} & Alexnet & 61.5 & -\\
        
        \citep{Twinanda2016Single-2016b} & Alexnet & 52.5 & -\\
    \end{tabular}
    \label{comparison}
\end{table*}
\section{Conclusion and Future Direction}
\label{conclusion}
The observation by surgical residents of the usage of specific surgical instruments and the duration of their usage in laparoscopic procedures gives great insight into how the surgery is performed. While identifying the tools in a recorded video of surgery is a trivial albeit tedious task for an average human, there are certain challenges in detecting the tools using computer vision algorithms. In order to tackle these challenges, in this paper, we proposed a novel deep learning system called LapTool-Net, for automatically detecting the presence of surgical tools in every frame of a laparoscopic video. The main feature of the proposed RCNN model is the context-awareness, i.e. the model learns the short-term and long-term patterns of the tools usages by utilizing the correlation between the usage of the tools with each other and, with the surgical steps, which follow a specific order. To achieve this goal, an LP-based model is used as a decision layer for the ML classifier and the training is performed in an end-to-end fashion. The advantage of this paradigm over direct LP classifier is that the training can be accomplished with a smaller dataset, due to having fewer classes and avoiding learning separate (and probably not useful) patterns for the superclasses. Furthermore, the order of occurrence of the tools is extracted through training a bi-RNN with the final prediction of a trained RCNN model. To overcome the high imbalance in the occurrence of the tools, we used under-sampling based on the tools’ combinations and the LP model. In addition to having a balanced dataset, the high under-sampling rate reduces the generalization error by avoiding overfitting, which is the main challenge in tool detection, due to the high correlation among videos frames. Our method outperformed all previously published results on M2CAI dataset, while using less than 1\% of the total frames in the training set.

While our model is designed based on the previous knowledge of the cholecystectomy procedure, it doesn't require any domain-specific knowledge from experts and can be effectively applied to any video captured from laparoscopic or even other forms of surgeries. Also, the relatively small dataset after under-sampling suggests that the labeling process can be accomplished faster by using fewer frames (e.g. one frame every 5 seconds). Moreover, the simple architecture of the proposed LP-based classifier makes it easy to use it with other proposed models such as \citep{AlHajj2018MonitoringNetworks} and \citep{Hu2017AGNet:Detection}, or with weakly supervised models [38] to localize the tools in the frames. Moreover, the offline design can be useful in generating a summary report, assessment and procedure rating etc. Also, the proposed model in online mode has a processing time of less than 0.01 seconds/frame, which makes it suitable for real-time applications such as feedback generation during surgery.

We plan on implementing a few ways to improve the performance of the proposed model. Firstly, the CNN can be replaced by a deeper and more accurate model. In particular, we will use the Inception-Resnet-V2 \citep{Szegedy2017Inception-v4Learning} for the CNN and the cholec80 dataset \citep{Twinanda2017EndoNet:Videosb} for training. Secondly, since the RNN doesn't extract the unique motion features of the tools, it can be replaced by a 3D CNN. The other way to improve the results is to choose better hyper-parameters, especially the class weights for balancing.

In future, we will investigate applying the findings in this paper to designing a semi-supervised learning based model \citep{Cheplygina2019Not-so-supervised:Analysis}, using only a fraction of the videos being labeled.

\section*{Acknowledgments}
This work was supported by a Joseph Seeger Surgical Foundation award from the Baylor University Medical Center at Dallas.

The authors would like to thank NVIDIA Inc. for donating the TITAN XP GPU through the GPU grant program.

\bibliographystyle{plainnat}

\end{document}